\documentclass{article}
\usepackage[acronym,smallcaps,nowarn,section,nogroupskip,nonumberlist]{glossaries}

\glsdisablehyper %

\newacronym{VAE}{vae}{variational auto-encoder}
\newacronym{IS}{is}{importance sampling}
\newacronym{IWAE}{iwae}{importance-weighted auto-encoder}
\newacronym{FIVO}{fivo}{Filtering Variational Objectives}
\newacronym{SMC}{smc}{sequential Monte Carlo}
\newacronym{SSM}{ssm}{state-space model}
\newacronym{SGA}{sga}{stochastic gradient ascent}
\newacronym{SGD}{sgd}{stochastic gradient descent}
\newacronym{ELBO}{elbo}{evidence lower bound}
\newacronym{KL}{kl}{Kullback-Leibler}
\newacronym{LSTM}{lstm}{long short-term memory}
\newacronym{CNN}{cnn}{convolutional neural network}
\newacronym{MML}{mml}{maximum marginal likelihood}
\newacronym{NVIL}{nvil}{Neural Variational Inference and Learning}
\newacronym{VIMCO}{vimco}{Variational inference for Monte Carlo objectives}

\newacronym{REINFORCE}{reinforce}{REINFORCE}
\glsunset{REINFORCE}
\newacronym{RL}{rl}{reinforcement learning}

\newacronym{ADAM}{adam}{ADAM}
\glsunset{ADAM}
\newacronym{RMSprop}{rmsprop}{RMSprop}
\glsunset{RMSprop}

\newacronym{GAN}{gan}{generative adversarial network}
\newacronym{GRU}{gru}{gated recurrent unit}
\newacronym{MLP}{mlp}{multilayer perceptron}
\newacronym{MC}{mc}{Monte Carlo}
\newacronym{mibp}{mIBP}{Markov Indian Buffer Process}
\newacronym{VRNN}{vrnn}{Variational Recurrent Neural Network}
\newacronym{LGSSM}{lgssm}{linear Gaussian state space model}
\newacronym[firstplural=recurrent neural networks, plural=RNNs]{RNN}{rnn}{recurrent neural network}

\newacronym{ELU}{elu}{exponential linear unit}

\newacronym{MNIST}{mnist}{mnist}
\glsunset{MNIST}

\newacronym{NMS}{nms}{non-maximum suppression}
\newacronym{DSPN}{dspn}{Deep Set Prediction Network}
\newacronym{CDSPN}{c-dspn}{size-conditioned \textsc{dspn}}
\newacronym{TSPN}{tspn}{Transformer Set Prediction Network}

\newacronym{SETMNIST}{set-mnist}{\textsc{mnist}-as-point-clouds}
\newacronym{CLEVR}{clevr}{clevr}
\glsunset{CLEVR}

\newacronym{IOU}{iou}{intersection-over-union}
\usepackage[utf8]{inputenc} %
\usepackage[T1]{fontenc}    %
\usepackage{url}            %
\usepackage{booktabs}       %
\usepackage{amsfonts}       %
\usepackage{nicefrac}       %
\usepackage{microtype}      %
\usepackage{appendix}
\usepackage{amsmath}
\usepackage{lipsum}
\usepackage{paralist}
\usepackage{stfloats} %
\usepackage{mathtools} %

\usepackage{todonotes}
\usepackage{subcaption}
\usepackage{bm} %
\usepackage{enumitem} %

\usepackage{mlmacros} %

\presetkeys{todonotes}{%
  backgroundcolor=blue!10!white,
  linecolor=blue!10!white,
  bordercolor=blue!10!white
}{}

\usepackage[skip=5pt,font=small]{caption}
\usepackage{subcaption}

\usepackage{titlesec} %

\variables{l,e,s,x,y,z,b,k,g,y,n,w,D,P,S}
\variables[mean]{\mu}
\variables[std]{\sigma}

\probdists{p,q}

\definecolor{pink}{HTML}{ff00ff}
\definecolor{orange}{HTML}{ff9900}
\definecolor{blue}{HTML}{0000ff}

\usepackage{hyperref}

\usepackage[accepted]{ool2020}

\usepackage[nameinlink,capitalise]{cleveref} %

\icmltitlerunning{Conditional Set Generation with Transformers}
\begin{document}

\twocolumn[
\icmltitle{Conditional Set Generation with Transformers}

\begin{icmlauthorlist}
\icmlauthor{Adam R. Kosiorek}{dm}
\icmlauthor{Hyunjik Kim}{dm}
\icmlauthor{Danilo J. Rezende}{dm}
\end{icmlauthorlist}

\icmlaffiliation{dm}{Deepmind, London, UK}

\icmlcorrespondingauthor{Adam R. Kosiorek}{adamrk@google.com}

\icmlkeywords{Machine Learning, ICML}

\vskip 0.3in
]

\printAffiliationsAndNotice{}  %

\begin{abstract}
A set is an unordered collection of unique elements---and yet many machine learning models that generate sets impose an implicit or explicit ordering.
Since model performance can depend on the choice of ordering, \textit{any} particular ordering can lead to sub-optimal results.
An alternative solution is to use a permutation-equivariant set generator, which does not specify an ordering.
An example of such a generator is the \gls{DSPN}.
We introduce the \glsreset{TSPN}\gls{TSPN}, a flexible permutation-equivariant model for set prediction based on the transformer, that builds upon and outperforms \gls{DSPN} in the quality of predicted set elements and in the accuracy of their predicted sizes.
We test our model on \gls{SETMNIST} for point-cloud generation and on \gls{CLEVR} for object detection.
\end{abstract}

\section{Introduction}
\label{sec:intro}
\begin{figure*}
    \centering
    \begin{minipage}[t]{0.49\linewidth}
        \centering
        \includegraphics[height=5cm]{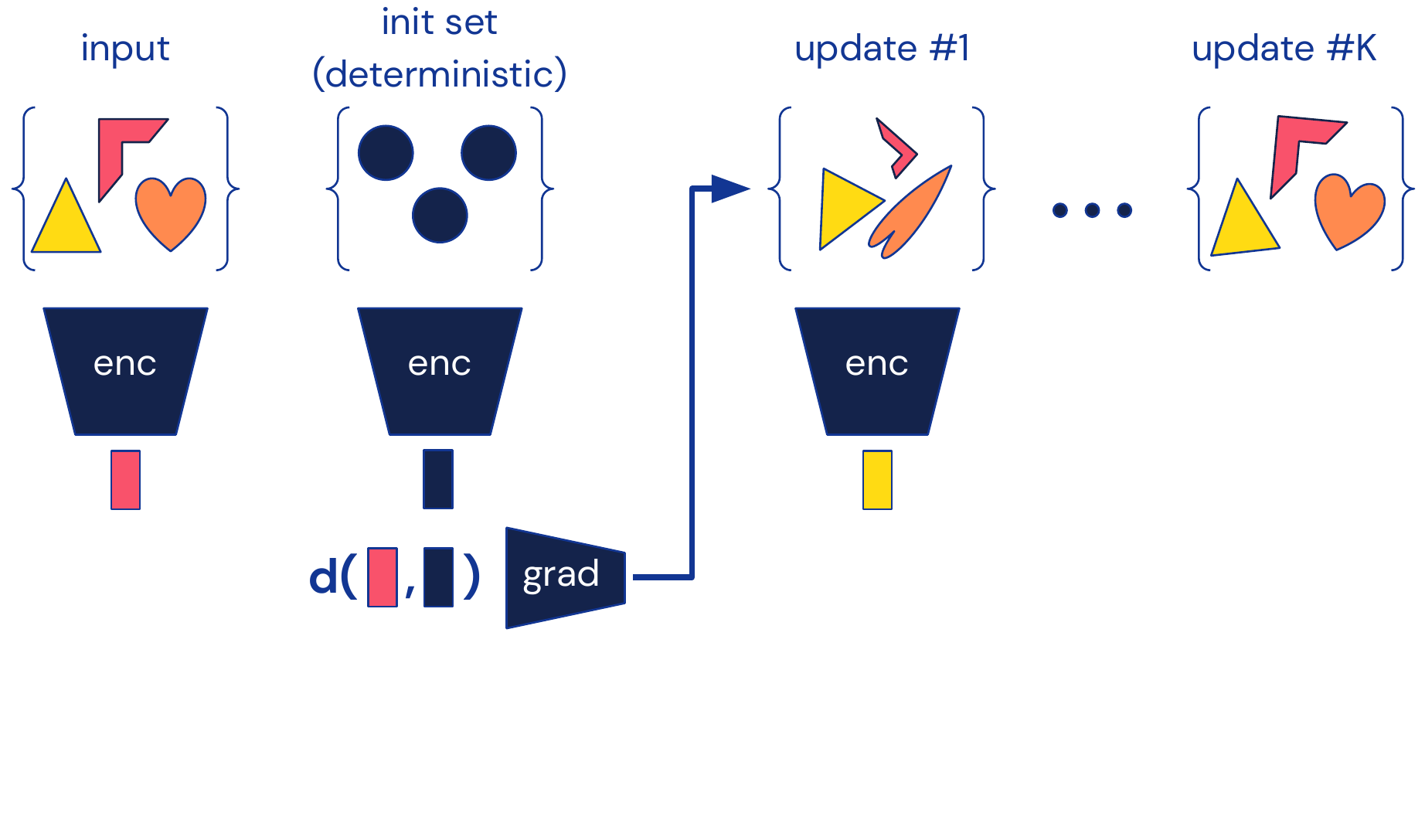}
        \vspace{-4.5em}
        \caption{
            \glsreset{DSPN}\Gls{DSPN} starts with a deterministic set that is gradually changed into the desired prediction with gradient descent on a learned loss function.
        }
        \label{fig:dspn}
    \end{minipage}
    \hfill
    \begin{minipage}[t]{0.49\linewidth}
        \centering
        \includegraphics[height=5cm]{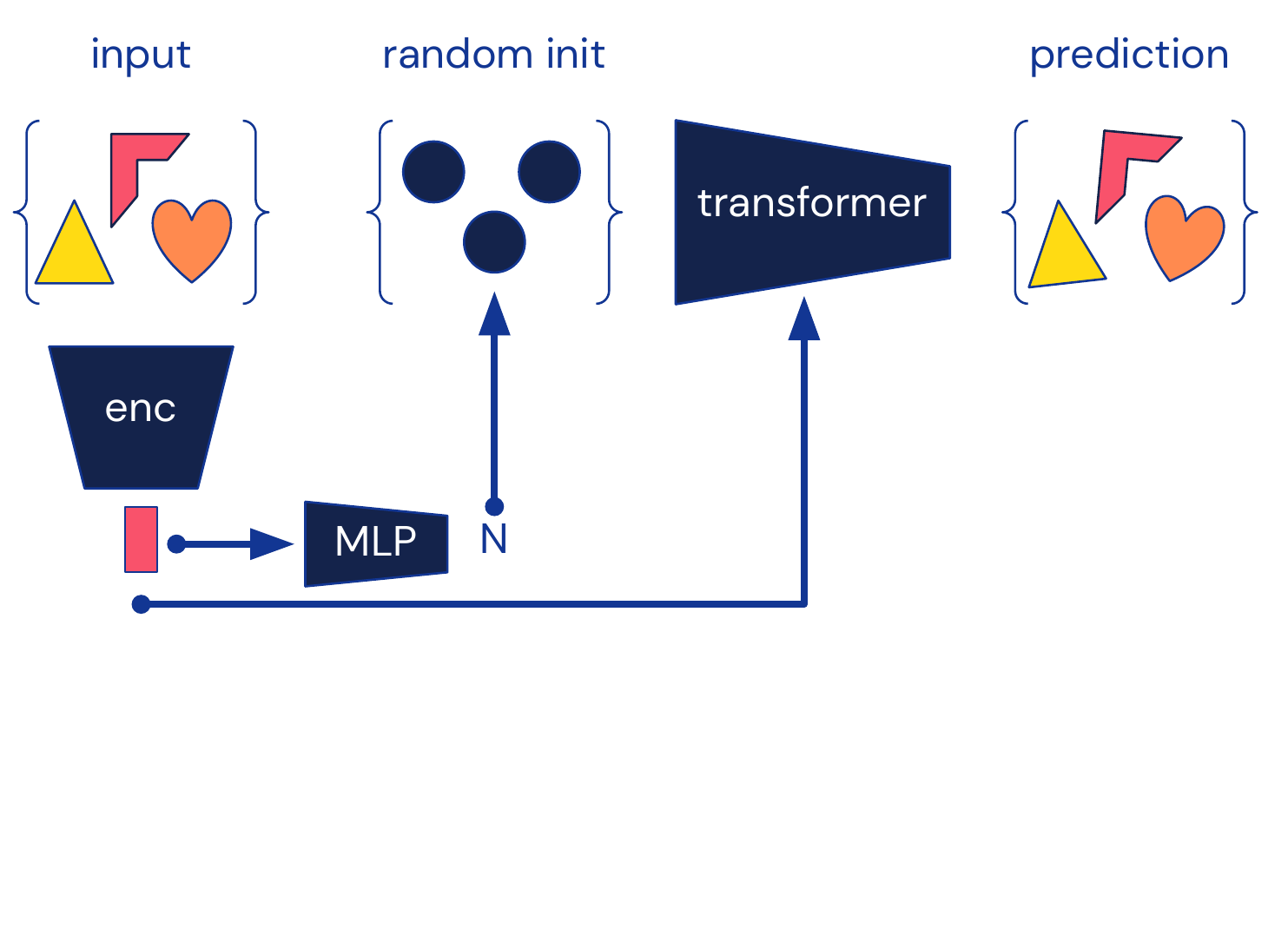}
        \vspace{-5.5em}
        \caption{
            Our \glsreset{TSPN}\gls{TSPN} extends \gls{DSPN} to use more expressive permutation-equivariant set transformations based on the Transformer \cite{Vaswani2017transformer}.
            Additionally, \gls{TSPN} explicitly learns the set cardinality, which allows it to generalize to much larger sets.
        }
        \label{fig:transformer}
    \end{minipage}
\end{figure*}

It is natural to reason about a group of objects as a set. 
Therefore many machine learning tasks involving predicting objects or their properties can be cast as a set prediction problem.
These predictions are usually conditioned on some input feature that can take the form of a vector, a matrix or a set.
Some examples include predicting future states for a group of molecules in a simulation \cite{No2020molecular}, object detection from images \cite{Carion2020detr} and generating correlated samples for sequential Monte Carlo in object tracking \cite{Zhu2020diffresampling,Neiswanger2014dependent}.
Elements of a set are unordered, which brings about two challenges that set prediction faces.
First, the model must be permutation-equivariant; that is, the generation of a particular permutation of the set elements must be equally probable to any other permutation.
Second, training a generative model for sets typically involves comparing a predicted set against a ground-truth set. 
Since the result of this comparison should not depend on the permutation of the elements of either set, the loss function used for training must be permutation-invariant.
While it is possible to create a set model that violates either or both of these requirements, such a model has to \emph{learn} to meet them, which is likely to result in lower performance.

Permutation equivariance imposes a constraint on the structure of the model \cite{Bloemreddy2018exchangable,Bloemreddy2019probabilistic}, and therefore sets are often treated as an ordered collection of items, which allows using standard machine learning models.
For example assuming that a set has a fixed size, we can treat it as a tensor and turn set-prediction into multivariate regression \cite{Achlioptas2018aae}.
If the ordering is fixed but the size is not, we can treat set prediction as a sequence prediction problem \cite{Vinyals2016seq2set}.
Both approaches require using permutation-invariant loss functions to allow the model to learn a deterministic ordering policy \cite{Eslami2016air}.
However imposing such an ordering can lead to a pathology that is commonly referred to as the \textit{responsibility problem} \citep{Zhang2019dspn, Zhang2020fspool}; there exist points in the output set space where a small change in set space (as measured by a set loss) requires a large change in the generative model's output. This can lead to sub-optimal performance, as shown in \citet{Zhang2020fspool}. 
Some approaches choose to learn the ordering of set elements \cite{Rezatofighi2018dpspn}, but this also suffers the same problem as well as adding further complexity to the set prediction problem.

Recently, \citet{Zhang2019dspn} introduced the \glsreset{DSPN}\gls{DSPN}---a model that generates sets in a permutation-equivariant manner using permutation-invariant loss functions.
\Gls{DSPN} relies on the observation that the gradient of a permutation-invariant function is equivariant with respect to the permutation of its inputs, also noticed by \citet{Papamakarios2019flowreview}.
\Gls{DSPN} uses this insight to generate a set by gradient-descent on a learned loss function with respect to an initially-guessed set.
\Gls{DSPN} has several limitations, however.
The functional form of the update step is limited, as the gradient information is only used to tanslate the set elements.
This, in turn, means that the method can be computationally costly: not only is the backward pass expensive, but many such passes might be needed to arrive at an accurate prediction.

In this paper we develop the \glsreset{TSPN}\gls{TSPN}, where we replace the gradient-based updates of \gls{DSPN} with a Transformer \cite{Vaswani2017transformer}, that is also permutation-equivariant and learns to jointly update the elements of the initial set.
We make the following contributions:
\begin{itemize}[leftmargin=*]
    \setlength\itemsep{-.25em}
    \item We show that the set-cardinality learning method of \citet{Zhang2019dspn} is prone to falling into local minima.
    \item We thus introduce an alternative, principled method for learning set cardinality.
    \item We learn a distribution over the elements of the initial set (as opposed to \gls{DSPN} that learns a fixed initial set).
    This allows one to directly generate sets with cardinality determined by the model (by sampling the correct number of points), and to dynamically change the size of the generated sets as test time.
    \item We demonstrate that \gls{TSPN} outperforms \gls{DSPN} on conditional point-cloud generation and object-detection tasks. 
\end{itemize}
We show that our model is not only more expressive than the \gls{DSPN}, but can also generalize at test-time to sets of vastly different cardinality than the sets encountered during training.
We evaluate our model on auto-encoding \gls{SETMNIST} \cite{Lecun2010mnist} and on object detection on \gls{CLEVR} \cite{Johnson2017clevr}.
We now proceed to describe \gls{DSPN} and our method \gls{TSPN} in detail, followed by experiments.

\section{Permutation-Equivariant Set Generation}
\label{sec:background}

\subsection{Permutation-Equivariant Generator}

The \glsreset{DSPN}\gls{DSPN} iteratively transforms an initial set into the final prediction, and the transformation is conditioned on some input.
That is, given a conditioning $\by \in \RR^{d_y}$ and an initial set $\bx \coloneqq \set{\bx_i^0}_{i=1}^N$ of $N$ points in $\RR^d$, \gls{DSPN} iteratively applies a permutation-equivariant $f$ to transform this initial set into the final prediction $\bx^K \coloneqq \set{\bx_i^K}_{i=1}^N$ over $K$ iterations:
\begin{equation}
    \set{\bx_i^K}_{i=1}^N = f^K\left(\set{\bx_i^0}_{i=1}^N, \by\right)\,.%
\end{equation}
In \gls{DSPN}, the points $\bx_i^0$ are initialised randomly as model parameters and learned, hence the model assumes fixed set cardinality.
For handling variable set sizes, each element of the predicted set is augmented with a \textit{presence} variable $p_i \in \interval{0, 1}$, which are transformed by $f$ along with the $x_i$ and then thresholded to give the final prediction.
The ground-truth sets are padded with all-zero vectors to the same size.
Note, however, that this mechanism does not allow to extrapolate to set sizes beyond the maximum size encountered in training.

\Gls{DSPN} employs a permutation-invariant set encoder (e.g.\ deep-sets \cite{Zaheer2017deepset}, relation networks \cite{Santoro2017relnet}) to produce a set embedding, and updates the initial prediction using the gradients of an embedding loss, arriving at a permutation-equivariant final prediction.

This leads to the following set-generation procedure.
Given the input embedding $\bm{h} = \operatorname{\mathrm{input\_encoder}}(\by)$, an initial set\footnote{We drop the explicit indication of cardinality to avoid clutter.} $\bx^0$, and a permutation-invariant $\mathrm{set\_encoder}$, we can arrive at the final prediction by performing gradient descent,
\begin{align}
    \widehat{\bm{h}}^k &= \operatorname{\mathrm{set\_encoder}} \left( \bx^k \right)\,,\\
    \bx^{k} = \bx^{k-1} - \lambda &\nabla_\set{\bx} \operatorname{d}(\bm{h}, \widehat{\bm{h}}^{k-1}) \text{ for } k=1,\ldots,K\,, %
\end{align}
with step size $\lambda~\in~\RR_+$, distance function $d$ and number of iterations $K$. 
The final prediction is $\bx^K$.
This set-generator can be used as a decoder of an autoencoding framework (with a permutation-invariant $\mathrm{input\_encoder}$) as well as for any conditional set-prediction task (\eg in object-detection where $\by$ is an image and $\mathrm{input\_encoder}$ is a \textsc{CNN}). 

\subsection{Permutation-Invariant Loss} \label{sec:loss}

The model is trained using a set loss, i.e.\ a permutation-invariant loss. Common choices are the Chamfer loss or the Hungarian loss:
\begin{align}
    \loss[\mathrm{cham}](A, B) &= \sum_{a \in A} \min_{b \in B} \operatorname{d}(a, b) + \sum_{b \in B} \min_{a \in A} \operatorname{d}(a, b) \label{eq:chamfer}\\
    \loss[\mathrm{hung}](A, B) &= \min_{\pi \in P} \sum_{a_i \in A} \operatorname{d}(a_i, b_{\operatorname{\pi}(i)}) \label{eq:hungarian}\,,
\end{align}
where $\pi$ is a permutation in the space of all possible permutations $P$ and $d$ can be any distance or loss function defined on pairs of set points.
Note that the computational complexity of the Chamfer loss is $O(N^2)$ for sets of size $N$, whereas the Hungarian loss is $O(N^3)$---it uses the Hungarian algorithm to compute the optimal permutation, whose complexity is $\mathcal{O}(N^3)$ \cite{Bayati2008mwn}.
Hence the Chamfer loss is suitable for larger sets, and the Hungarian loss for smaller sets.
For $d$, \citet{Zhang2019dspn} use the Huber loss defined as $\operatorname{d}(\bm{a}, \bm{b}) = \sum_i \min\left(\nicefrac{1}{2}(a_i-b_i)^2, |a_i-b_i| - \nicefrac{1}{2}\right)$\,.

Recall that in the implementation of \gls{DSPN}, the ground truth set is padded with zero vectors so that all sets have the same size. Padding a set $A$ to a fixed size with constant elements $\widehat{a}$ turns it into a \textit{multiset} $\widehat{A}$. A Multiset is a set that contains repeated elements, that can be represented by an augmented set where each unique element is paired with its multiplicity.
If we use a multiset in its default form (\ie with repeated elements) as the ground-truth in the Chamfer loss, then it is enough for the model to predict a set $B$ containing exactly one element $b_i=\widehat{a}$ equal to the repeated element of the set $\widehat{A}$ in order to account for all its repetitions in the first term of \Cref{eq:chamfer}.
The remaining superfluous elements $b_j \in B$ predicted by the model can match \textit{any} other element in $A$ without increasing the second term of \Cref{eq:chamfer}.
This implies that padding a ground-truth set of size $N$ with $M$ constant elements creates $\binom{N+1}{M-1}$ predictions that are all optimal and hence indistinguishable under the Chamfer loss.
These predictions have a set size that varies from $N$ to $N+M-1$, and hence the model is likely to fail to learn the correct set cardinality---an effect clearly visible in our experiments, \cf \Cref{sec:set_mnist} and \Cref{tab:set_mnist}.

\Gls{DSPN} uses an additional regularization term {\small $\loss[\mathrm{repr}]{\bm{h}, \widehat{\bm{h}}} = \operatorname{d}(\bm{h}, \widehat{\bm{h}})$} defined as the distance between the model's input features and the encoding of the generated set, which improves the performance of the internal optimization loop.

\section{Size-Conditioned Set Generation with Transformers}
\label{sec:method}
We follow \gls{DSPN} in that we generate sets by transforming an initial set.
However, we employ Transformers \cite{Vaswani2017transformer,Lee2019settransformer} as a learnable set-transformation, which can readily account for any interactions between the set elements.
This change implies that the elements of the initial set can have a different (higher) dimensionality to the elements of the final predicted set, increasing the flexibility of the model.
Moreover, instead of concatenating a presence variable to each set point, \gls{TSPN} uses a \gls{MLP} to directly predict the number of points from the input embedding $\bm{h}$, which then chooses the size $N$ of the initial set $\set{\bx}_{i=1}^N$.
Given this size, we must generate the elements of the initial set such that the final predicted set is permutation-equivariant. One solution is to sample the intial set from a permutation-invariant distribution \eg \textit{iid} samples from a fixed distribution 
, for which we choose $\gauss{\alpha\bm{1}, \operatorname{diag}(\beta\bm{1})}$ with learnable $\alpha \in \RR$, $\beta \in \RR_+$.
This makes the model stochastic, but also allows one to choose an arbitrary size for the generated set, even to sizes that lie outside the range of set sizes encountered in training.
Formally, let $\bm{h}$ be the input embedding; the generating process reads as follows:
\begin{align}
    N &= \operatorname{\textsc{mlp}}(\bm{h})\,,\\
    \bx_i \sim&~\gauss{\alpha\bm{1}, \operatorname{diag}(\beta\bm{1})}, \quad i=1,\ldots,N\,,\\
    \widehat{\bx}_i &= \operatorname{\mathrm{concatenate}}(\bx_i, \by)\,,\\
    \set{\bx} &= \operatorname{\mathrm{transformer}}\left(\set{\widehat{\bx}}\right)\,.%
\end{align}
While training \gls{TSPN}, we use the ground-truth set-cardinality to instantiate the initial set, and we separately train the \gls{MLP} by minimizing categorical cross-entropy with the ground-truth set sizes.
The cardinality-\gls{MLP} is used only at test-time.
Note that, in contrast to \gls{DSPN}, \gls{TSPN} does not require any additional regularization terms applied to its representations.

We describe our work in relation to \gls{DSPN} \cite{Zhang2019dspn}, but we note that there are concurrent works that share the ideas presented here.
Both \textsc{detr}~\cite{Carion2020detr} and Slot Attention~\cite{Locatello2020slot} use a variant of the transformer \cite{Vaswani2017transformer} for predicting a set of object properties.
\textsc{Detr} uses an object-specific query initialization and is, therefore, not equivariant to permutations, similarly to \cite{Zhang2019dspn}.
Slot Attention is perhaps the most similar to our work---transforms a randomly sampled point-cloud, same as \gls{TSPN}, but it uses attention normalized along the query axis instead the key axis.
\section{Experiments}
\label{sec:experiments}
We apply \gls{TSPN} and our implementations of \gls{DSPN} and a \gls{CDSPN} to two tasks: point-cloud prediction on \gls{SETMNIST} and object detection on \gls{CLEVR}.
Point-cloud prediction is an autoencoding task, where we use a set encoder to produce vector embeddings of point clouds of varying sizes, and explore the use of the above set prediction networks for reconstructing the point clouds conditioned on these embeddings.
Object detection, instead, requires predicting sets of bounding boxes conditioned on image features.
While generating point-clouds requires predicting large numbers of points, which are often assumed to be conditionally independent of each other, detecting objects typically requires generating much smaller sets.
Due to possible overlaps between neighbouring bounding boxes, however, it is essential to take relations between different objects into account.
We implemented all models in \textsc{jax} \cite{Jax2018github} and \textsc{haiku} \cite{Haiku2020github} and run experiments on a single Tesla V100 GPU. 
Models are optimized with \textsc{adam} \cite{Kingma2015adam} with default $\beta_1, \beta_2$ parameters.
We used batch size $=32$ and trained all models for $100$ epochs.

\subsection{Point-Cloud Generation on \textsc{Set-Mnist}}
\label{sec:set_mnist}

    \begin{figure}
        \centering
        \includegraphics[width=\linewidth]{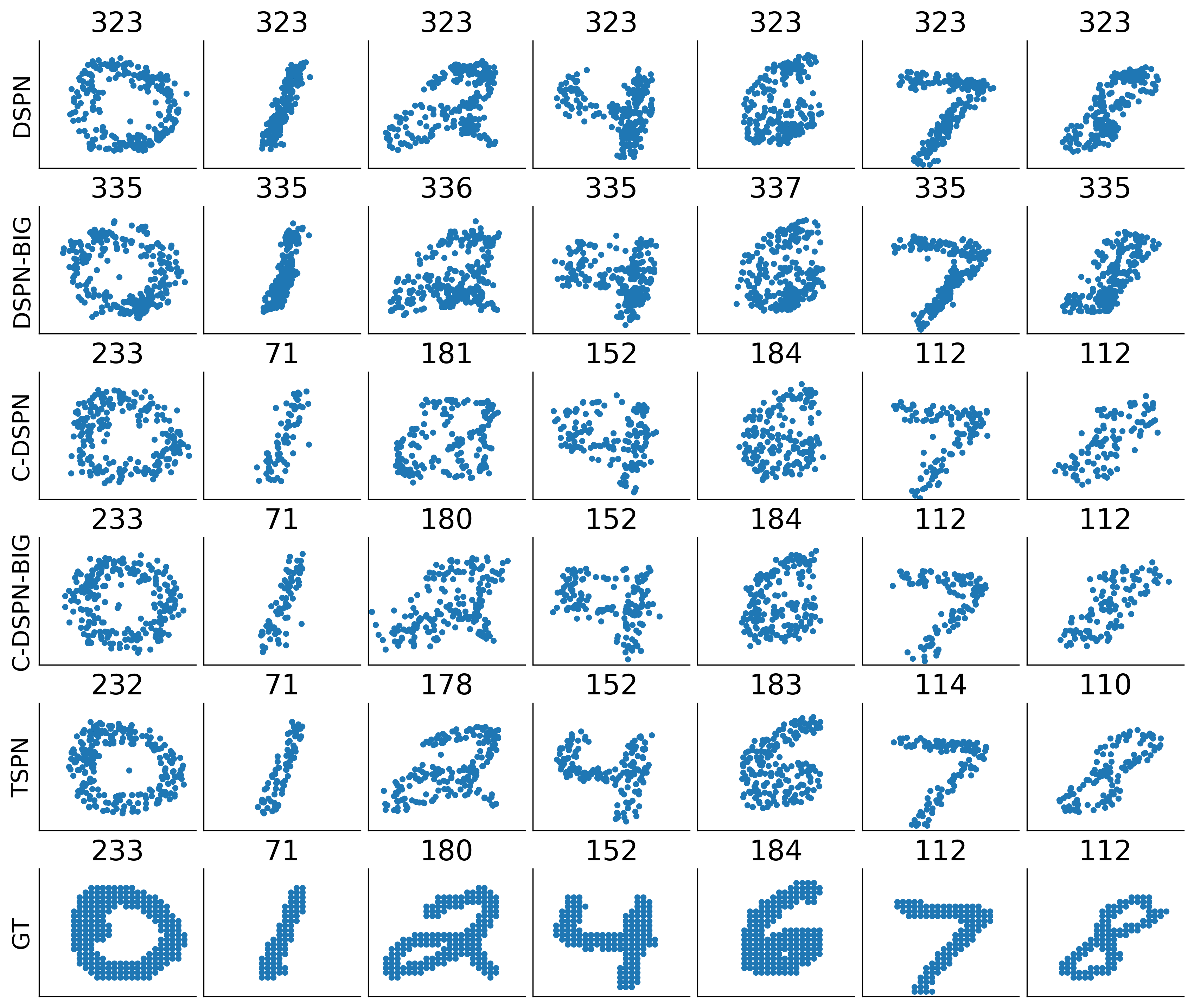}
        \caption{Set reconstructions (rows 1-5) and inputs (row 6) from different models, annotated with the number of points (above). 
        \Gls{DSPN} fails to learn set cardinality, see \Cref{sec:loss} for an explanation of why this happens.
        \Gls{CDSPN} learns the cardinality well, but produces reconstructions of lower fidelity than \gls{TSPN}.
        }
        \label{fig:mnist}
    \end{figure}
    
    \begin{table}
        \centering
    
        \begin{tabular}{c|c|c}
            \toprule
            Model & Chamfer $\left[10^{-5}\right]$ $\downarrow$ & Set Size \textsc{Rmse} $\downarrow$\\
            \midrule
            \textsc{dspn} & $8.2 \pm 0.66$ & $165 \pm 54.7$\\
            \textsc{dspn-big} & $8.5 \pm 1.83$ & $182 \pm 29.2$\\
            \textsc{c-dspn} & $8.92 \pm 0.78 $ & \bm{$0.3 \pm 0.13$}\\
            \textsc{c-dspn-big} & $9.63 \pm 1.17$ & \bm{$0.3 \pm 0.09$}\\
            \textsc{tspn}& \bm{$5.42 \pm 0.15$} & $0.8 \pm 0.08$\\
            \bottomrule
        \end{tabular}
    
        \caption{
            \Gls{SETMNIST} results, averaged over 5 runs.
            \textsc{Rmse} is root-mean-square error.
            Note that our \gls{DSPN} results match the Chamfer loss reported in \cite{Zhang2019dspn}.
            \textsc{C-dspn} stands for the \gls{DSPN} conditioned on the number of points.
            The \textsc{-big} variants use wider layers with the same number of parameters as \gls{TSPN}.
            For conditional models, sets were generated with the number of points inferred by the model.
        }
        \label{tab:set_mnist}
    \end{table}

We convert \gls{MNIST} into point clouds by thresholding pixel values with the mean value in the training set; then normalize coordinates of remaining points to be in $\interval{0, 1}$. 

We use the same hyperparameters for \gls{DSPN} as reported in \cite{Zhang2019dspn}.
\gls{CDSPN} uses the same settings, except instead of presence variables $p_i$ it uses an \gls{MLP} with one hidden layer of size $128$ to predict the set size, as used in \gls{TSPN}.
\Gls{TSPN} uses a three-layer Transformer, with parameters shared between layers\footnote{This gives 383k parameters compared to 190k for \gls{DSPN}, which shares parameters between its encoder and gradient-based decoder. Not sharing parameters between layers does not improve results, but significantly increases parameter count. Sharing parameters and increasing layer width to match the number of parameters does not increase performance, either.}.
Each layer has $256$ hidden units and $4$ attention heads.
The initial set is two-dimensional (same as the output), but it is linearly-projected into $256$-dimensional space.
The outputs of the transformer layers are kept at $256$ dimensions, too, and we project them back into $2$ dimensions after the last layer.
\Gls{TSPN} uses the same input encoder as (\textsc{c}-)\gls{DSPN}, that is a two-layer \textsc{deepset} with \textsc{fspool} pooling layer \cite{Zhang2020fspool}.
All models are trained using the Chamfer loss with learning rate~$=10^{-3}$.

\Cref{tab:set_mnist} shows quantitative results.
Conditioning on the set size does not improve the Chamfer loss for \gls{CDSPN} but it does significantly improve the accuracy of set-size prediction.
Further, replacing the decoder with the Transformer (\Gls{TSPN}) leads to a significant loss reduction with respect to \gls{DSPN}.
\Cref{fig:mnist} shows inputs set and model reconstructions. 
Note that in our experiments, \gls{DSPN} almost always predicts the same number of points ($323$), which is close to the maximum number of points we used (here, $342$).
Interestingly, \gls{TSPN} performs very well while extrapolating to much bigger sets than the ones encountered in training: \Cref{fig:tspn_mnist_extrapolation} in \Cref{sec:mnist_app} shows reconstructions, where we manually change the desired set size up to $1000$ points.
This is in contrast to \gls{CDSPN}, whose performance decreases significantly when we require it to generate a set whose size differs only slightly from the input, \cf \Cref{fig:dspn_mnist_extrapolation} in \Cref{sec:mnist_app}.
We conjecture that this is caused by how \textsc{fspool} handles sets of different sizes, which leads to incompatibility between embeddings of sets of different cardinality.
This, in turn, causes the distances in the latent space to be ill-defined, and breaks the internal gradient-based optimization. 

\subsection{Object Detection on \textsc{Clevr}}

      \begin{figure}
        \centering
        \includegraphics[width=\linewidth]{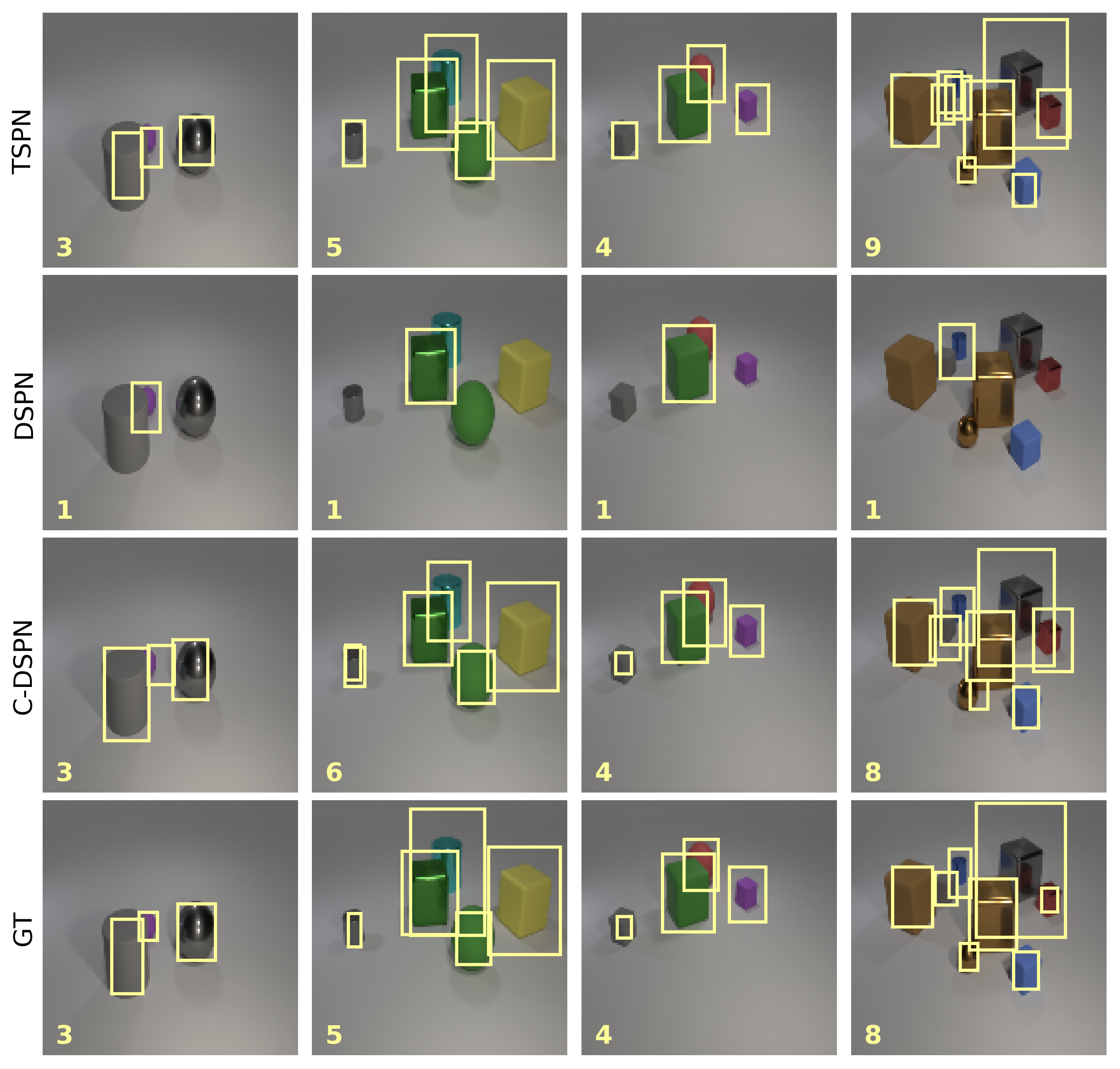}
        \caption{
            Object detection on \gls{CLEVR}, with the number of detected objects in the lower-left corner of each image.
            \Gls{DSPN} fails to detect the correct number of objects.
            The ground-truth bounding boxes are approximate.
        }
        \label{fig:clevr}
        \vspace{-1em}
    \end{figure}

       \begin{table*}
        \centering
    
        \begin{tabular}{c|c|c|c|c|c|c|c}
            \toprule
            Model & $\textsc{ap}_{50}$ & $\textsc{ap}_{95}$ & $\textsc{ap}_{95}$ & $\textsc{ap}_{98}$ & $\textsc{ap}_{99}$ & Set Size \textsc{Rmse} $\downarrow$ &  \#Params\\
            \midrule
            \textsc{dspn} & $67.7 \pm 5.49$ & $7.4 \pm 0.91$ & $0.6 \pm 0.10$ & $0.0 \pm 0.01$ & $0.0 \pm 0.00$ & $2.53 \pm 0.221$ & $0.3$\,M\\
            \textsc{c-dspn} & $71.6 \pm 3.40$ & $10.8 \pm 1.50$ & $0.9 \pm 0.21$ & $0.0 \pm 0.01$ & $0.0 \pm 0.00$ & $1.74 \pm 0.301$ & $0.3$\,M\\
            \textsc{tspn}& \bm{$81.2 \pm 1.03$} & \bm{$20.7 \pm 0.16$} & \bm{$3.0 \pm 0.20$} & \bm{$0.1 \pm 0.02$} & $0.0 \pm 0.00$ & \bm{$0.58 \pm 0.046$} & $1.9$\,M\\
            
            \bottomrule
        \end{tabular}
        \caption{
            \Gls{CLEVR} object detection results averaged over 5 runs.
            These results are not directly comparable to the ones reported in \citet{Zhang2019dspn}, since all our models were trained using Chamfer loss. We subtract the resnet parameters ($21.8$\,M) when reporting the number of model parameters.
        }
        \label{tab:clevr}
    \end{table*}
    
\Gls{CLEVR} images consist of up to 10 rendered objects on plain backgrounds.
Following \citet{Zhang2019dspn}, we use the \gls{CLEVR} dataset to test the efficacy of our models for object detection in a simple setting, which might pave the way for more advanced object-based inference in future works.
We use the same hyperparameters for \gls{CDSPN} as \citet{Zhang2019dspn}.
\Gls{TSPN} uses four layers with four attention heads and $256$ neurons each, without parameter sharing between layers.
We apply layer normalization \cite{Xiong2020Oonln} before attention as in \cite{Ba2016ln}.
The last transformer layer is followed by an \gls{MLP} with a single hidden layer of %
All models use a \textsc{resnet34} \cite{He2016resnet} as the input encoder, and are trained with the Chamfer loss; \gls{DSPN}-based models are trained for $200$ epochs with learning rate~$=3\times10^{-5}$; \gls{TSPN} uses learning rate~$=10^{-4}$ and is trained for $1200$ epochs.
Longer training for \gls{DSPN}-based models lead to overfititng and decreased validation performance.
Note that \gls{DSPN} in \cite{Zhang2019dspn} use the Hungarian loss (\Cref{eq:hungarian}), which leads to better results than using Chamfer.
We report the average precision scores at different thresholds and the set size root-mean-square error in \Cref{tab:clevr}.
Qualitative results are available in \Cref{fig:clevr}.
We see that \gls{TSPN} outperforms \gls{CDSPN} and produces bounding boxes that are better aligned with ground-truth, although these results are worse than the ones reported in \citet{Zhang2019dspn}---we expect improvements when using the Hungarian loss as well.
\section{Conclusions}
\label{sec:conclusion}
We introduced the \glsreset{TSPN}\gls{TSPN}---a transformer-based model for conditional set prediction.
\Gls{TSPN} infers the cardinality of the set, randomly samples an initial set of the desired size and applies a transformer to generate the final prediction.
Set prediction in \gls{TSPN} is permutation-equivariant, and the model can be applied to any set-prediction tasks.
Interesting directions include scaling the model to large-scale point-clouds and object detection (\eg similar to \citet{Carion2020detr}), as well as turning this model into a generative model in either the \textsc{vae} or \textsc{gan} framework.

\section*{Acknowledgements}
We would like to thank George Papamakarios, Karl Stelzner, Thomas Kipf, Teophane Weber and Yee Whye Teh for helpful discussions.

\bibliography{library}

\newpage
\appendix
\section{Set Size Extrapolation on \textsc{Set-Mnist}}
\label{sec:mnist_app}

\begin{figure*}[b]
    \centering
    \begin{minipage}[t]{.49\linewidth}
        \centering
        \includegraphics[width=\linewidth]{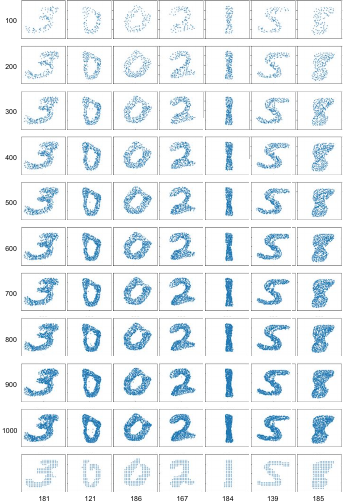}
        \caption{
            \Gls{TSPN} extrapolates to far bigger set sizes than encountered in training.
            Here, the model was trained with up to 342 points, and yet can generate sets of up to 1000 points.
            The bottom row contains the ground-truth annotated with the number of points at the bottom.
            This figure uses a smaller marker size than \Cref{fig:mnist}, hence ground-truth appears less dense.
        }
        \label{fig:tspn_mnist_extrapolation}
    \end{minipage}
    \hfill
    \begin{minipage}[t]{.49\linewidth}
        \centering
        \includegraphics[width=\linewidth]{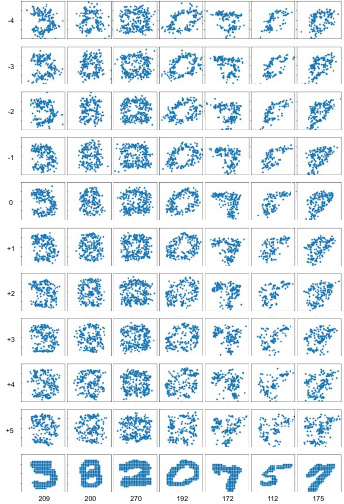}
        \caption{
            The size-conditional \gls{DSPN} fails to reconstruct sets when we slightly change the output set size.
            Here, we take the size of the reconstructed set to be within $\interval{-4, 5}$ points of the input size.
            As the size of the reconstruction strays from the original size, the reconstruction quality quickly deteriorates.
            The bottom row contains the input sets with their respective sizes.
        }
        \label{fig:dspn_mnist_extrapolation}
    \end{minipage}
\end{figure*}
\end{document}